\documentclass[11pt]{article}

\usepackage[T1]{fontenc}
\usepackage{setspace}
\usepackage{indentfirst}
\usepackage{amsmath}
\usepackage{abstract}
\usepackage[linesnumbered, vlined, ruled]{algorithm2e}
\usepackage{float}
\usepackage{footnote}
\usepackage{titling}
\usepackage{dialogue}
\usepackage{titlesec}
\usepackage{lipsum}
\usepackage{hyperref}
\usepackage{fancyhdr}
\usepackage{multicol}
\usepackage{graphicx}
\usepackage{caption}
\usepackage[table]{xcolor}
\usepackage{setspace}
\usepackage{authblk}
\usepackage{lipsum}
\usepackage[columnwise, modulo]{lineno}
\usepackage[margin=0.8in]{geometry}

\graphicspath{ {./resources/} }

\usepackage[backend=biber]{biblatex}
\DeclareCaptionType{equ}[Equation][List of equations]

\DeclareCaptionType{dia}[Dialogue][List of dialogues]

\newcommand{\theactualtitle}{\Large \textbf{Towards Automated Psychotherapy via Language Modeling}}
\title{\theactualtitle}
\author{
    Houjun Liu \\
    {The Nueva School \\ 131 E. 28th Ave., \\ San Mateo, California} \\ \smallskip
    \texttt{houliu@nuevaschool.org}}

\date{}

\newcommand{\specialcell}[2][c]{%
  \begin{tabular}[#1]{@{}c@{}}#2\end{tabular}}
 
\renewenvironment{abstract}
 {\small
  \begin{center}
  \bfseries \abstractname\vspace{-.5em}\vspace{0pt}
  \end{center}
  \list{}{%
    \setlength{\leftmargin}{20mm}
    \setlength{\rightmargin}{\leftmargin}%
  }%
  \item\relax}
 {\endlist}

 \newcommand{\ignore}[1]{}

 \titleformat{\subsubsection}[runin]
 {\bfseries}
 {}
 {0em}
 {}

\begin{document}
% DID YOU SET SPELL??????
%\onehalfspacing

\maketitle

\vspace{-3em}

\begin{abstract}
    In this experiment, a model was devised, trained, and evaluated to automate psychotherapist/client text conversations through the use of state-of-the-art, Seq2Seq Transformer-based Natural Language Generation (NLG) systems. Through training the model upon a mix of the Cornell Movie Dialogue Corpus for language understanding and an open-source, anonymized, and public licensed psychotherapeutic dataset, the model achieved statistically significant performance in published, standardized qualitative benchmarks against human-written validation data --- meeting or exceeding human-written responses' performance in $59.7\%$ and $67.1\%$ of the test set for two independent test methods respectively. Although the model cannot replace the work of psychotherapists entirely, its ability to synthesize human-appearing utterances for the majority of the test set serves as a promising step towards communizing and easing stigma at the psychotherapeutic point-of-care.

\end{abstract}

\begin{multicols}{2}
    %cool. right? \autocite{vaswash2017}

    \section{Introduction}
    Since the advent of the ELIZA ``natural language'' dialogue system --- a mock Rogerian psychologist --- in the 1960s, the field of natural language processing has grown significantly. Instead of using a predefined subset of rules manually written to interact with human agents, state-of-the-art dialogue systems now employ natural language understanding models (chief among which artificial neural networks) to probabilistically generate the appropriate corresponding responses.

    Until very recently, Recurrent Neural Networks (RNNs) have been the gold standard for natural language synthesis systems. Their recursive nature allows for sequential understanding of text; however, due to their large and densely connected topology, these networks often experiences training-time problem that hinders convergence. To assist in contextual understanding and convergence of RNNs, the ``attention'' mechanism was developed to create trainable context-gathering vectors that are used in conjunction with recurrent topologies.

    The proposal of the Transformer \autocite{vaswash2017}, however, sought to rely solely on the attention mechanism and removed traditional recurrent elements from natural language networks entirely. This network is able to achieve state-of-the-art performance without the usual difficulties of training a recurrent network.

    These new systems, although providing high-quality utterances once fully trained, require an immense set of data to be able to converge. Furthermore, although the networks are much easier to converge due to their reasonably shallow nature, they are still highly memory and resource intensive to train.

    Through the use of encoder-decoder style Transformer networks, this experiment proposes to replicate --- with NLG --- psychotherapist-client conversations. To address the dearth of high-quality psychotherapy utterances, a technique similar to ``transfer-training'' is employed --- by first training a base language understanding model using a gold-standard corpora, then proceeding to fit the model on a mix of the original dataset and a much smaller open-source psychotherapy dataset. Finally, the resulting network was analyzed based on a set of quality indexes (QIs) devised to benchmark the network against validation data with human-written responses.

    \section{Background}
    \subsection{Automated Psychotherapy}
    In the CDC's 2016 national survey, almost 10\% of physician visits have depression indicated on their clinical record \autocite{cdc2021}. Furthermore, according to the Pew Research Center, 70\% of teenagers believe that anxiety and depression is a major problem among their peers \autocite{horojul2019}.

    Although many efficient systems (e.g. hot-lines, community watch, etc.) have been put in place as ``emergency services'' for the prevention of suicide and self harm --- extreme potential outcomes of depression --- much less attention has been placed on the ease of access to immediate, free psychotherapy conversations in less extreme cases. Due both to the lack of adequate resources --- 47\% of Americans believe that their access to psychotherapies is limited, and 25\% reported that their access to such therapies will limit their access to daily necessities --- and the social stigma associated with psychotherapy --- 31\% of Americans have worried about peer judgement rising from access to psychotherapy services --- \autocite{nationa2018} the need for free, on-demand, and anonymous psychotherapies are more pressing than ever.

    Even though the need of trained psychologists is and will remain pressing, the development of online automated psychotherapy systems will reduce the load on trained professionals and --- with the combination of a sentiment analysis system --- will allow direct takeover by professionals when anomalies are detected.

    Attempts have been made for the development of automated psychotherapy. \autocite{althtim2016} Such algorithms are reliant upon message-to-message Hidden Markov Models to combine fragments of trained conversation together in response to user utterances. This work not only leverages the Markov assumption as used in previous work, but also uses novel natural language generation (NLG) systems to create more fluid and personalized conversations.

    \subsection{Transformer Models} \label{transformers}
    Ever since \textit{Vaswani, et al.} released the topology of the attention-based transformer, variants upon that network topology has set new benchmarks in the natural language synthesis field \autocite{vaswash2017}. As originally proposed, the Transformer network excels at machine translation tasks (scoring 28.4 BLEU --- 2 points above the state-of-the-art of the time) and could be generalized to tasks such as English constituency parsing (the parsing of sentence structure) with similarly high-performing results (achieving a F1 measure of 92.7\% in the semi-supervised training case.)

    %In recent years, further improvements upon the network has been made to perform different tasks with equal success: with the creation of GPT-2/3 setting benchmarks for machine text generation and BERT for masked language modeling and question-answering.

    This work operates under a Markovian assumption of the dataset; meaning, the model currently aim only to represent psychotherapy conversation as context-independent, pair-wise conversations. As such, the encoder-decoder topology designed for translation in the originally proposed Transformer is suitable for the task purposed here.

    \begin{figure}[H]
        \centering
        \includegraphics[width=0.4\textwidth]{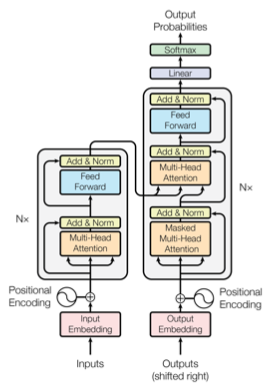}
        \caption{\small Transformer Topology, \textit{Vaswani, et al. \autocite{vaswash2017}}}
    \end{figure}

    Encoder-decoder transformer models are constructed with a generally typical encoder-decoder architecture for traditional neural nets, but embedded with not feed-forward but especially created attention-driven encoding/decoding blocks placed in sequence of each other.

    \subsubsection{Self-Attention} \label{selfattn}
    Self-attention modules are matrices that enables the calculation of the value of each token --- a fragment (commonly a word) in an utterance --- with respect to all other tokens in a phrase. Originally proposed to aid training of recurrent neural networks, it is discovered that these values can serve by themselves a model for language in a NLG system.

    Three trainable parameters are leveraged to calculate self-attention. Query $Q_i$ is the vector representation of the current token in question; key $K$ represents a matrix encoding the attention values of each possible query; and value $V$ represents the importance placed upon each position in the utterance. 

    Hence, the calculation of the ``self-attention'' of a token is simply a lookup process whereby the scaled, softmax output of $Q_i \cdot K$ (the query's value in the key) is crossed with $V$ (the importance of that query value.)

     \begin{figure}[H]
        \centering
        \includegraphics[width=0.2\textwidth]{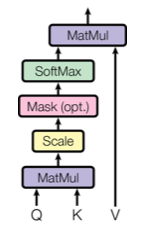}
        \caption{\small Scaled Dot-Product Attention, \textit{Vaswani, et al. \autocite{vaswash2017}}}
    \end{figure}

    In order to prevent oversized gradient updates, each product of $Q \cdot K$ (notably, using $Q$ and not $Q_i$ as attention could be processed with multiple queries at a time) is scaled by $\sqrt{D_k}$ --- dimension of the model key vectors $K$. Formally, $Attn(Q, K, V) = softmax(\frac{QK}{\sqrt{D_k}})V$. \autocite{vaswash2017}

    In practice, an attention module named ``Multi-Headed Attention'' is used. This is a form of attention whereby the second-dimension sum (concatenation) of the dot-product attention values attending to different queries is taken after being passed through a linear feed-forward neural network layer. In this manner, the network could attend to multiple tokens at once to recognize broader patterns in the utterance.

    \subsubsection{Positional Encoding} \label{posenc}
    One of the original drawbacks of the transformer architecture was, unlike its deeper recurrent network counterpart, it does not have inherent spacial awareness regarding the position of a token within a phrase. However, \textit{Vaswani, et al.} proposed the process of Positional Encoding, whereby for every token $i$ in a sequence, the value of

    \begin{equ}[H]
        \begin{equation}
            \begin{cases} 
                sin(\frac{pos}{10000}^{2i/d_{model}}) & 2i\\
                cos(\frac{pos}{10000}^{2i/d_{model}}) & 2i+1\\
            \end{cases}
        \end{equation}
        \caption{Positional Encoding}
    \end{equ}

    is added to the original input vector --- creating a sense of positional awareness by imposing a ``checkerboard''-style matrix oscillating between odd/even tokens over the whole input sequence. 

    Through this technique, the model gains heightened spacial awareness without the need to create a fully recurrent network.

    \subsubsection{Transformer Topology}
    The encoder-decoder transformer model is split into two major blocks: the ``encoder'' block --- originally designed for parsing source translation language and used here to parse client utterance --- and the ``decoder'' block --- designed originally to synthesize translation target language and here used to mimic psychotherapist utterance.

    In the network, client utterance input is supplied to the encoder block, and previous tokens of the currently constructed psychotherapist output is supplied to the decoder block. The output values of the encoder block is also passed to the decoder block in the manner mentioned below.

    The encoder block consists of an initial layer of embedding, followed by the process of positional encoding as mentioned in section \ref{posenc}, and finally into rounds of attention blocks.

    In each encoder attention block, a unit of multi-headed self-attention described in section \ref{selfattn} is connected to a skip-connection sum (correcting other attention blocks from ``below'' the network) and finally to a traditional layer-normalization layer \autocite{vaswash2017} designed to stabilize weights.

    Through this experiment, 4 such encoder blocks were placed together to form the encoder. Finally, a feed-forward and normalize layer caps the output of the encoder.

    The decoder section behaves in much the same way as the encoder. However, in the second attention layer of the decoder, the attention target is switched to the output of the encoder; that is, instead of decoder supplied $K$ and $V$ matrices as in the whole encoder and the first layer of decoder (creating self-attention), the values of $K$ and $V$ are supplied by the encoder output. In this manner, the decoding response will not only be based on previous tokens in the output/psychotherapist utterance, but also in the tokens of the input/client utterance.

    The final result of the decoder will be activated using the softmax activation function, and passed as a probabilistic map of possible next tokens. The token with the highest probabilistic confidence is taken as the next token written by the model.

    \subsubsection{Loss Function}
    The model was optimized for standard Cross-Entropy loss masked for pad characters (i.e. the generation of an incorrect token against padding does not count towards loss.)

    \section{Experimental Methods}
    \subsection{Data Handling}
    \subsubsection{Data Acquisition}
    There is a dearth of high-quality, insensitive, and anonymized psychotherapist/client conversations available for machine learning research. Indeed, the factors needed for the acquisition of such datasets are rather incongruous as most high-quality transcripts are heavily sealed under, if not HIPPA, other forms of privacy agreements.

    However, anonymous message boards occupied by professional psychotherapists provides a healthy source of pairwise client-therapist data that are both not protected by regulations such as HIPPA and professionally cleaned of personal information.

    Under the direction of Dr. Grin Lord and project co-founders Eric Ström and Phil Lee of online psychotherapy website \textit{counselchat.com} \autocite{counselc2021}, accredited professional responses to user-submitted questions on their website were gathered by a third party and released to the public under the MIT Permissive License.

    \subsubsection{Data Cleaning}
    As the original dataset is collected from a massively online chat system, it is not immediately suitable for sequence modeling. First, to reduce the vocabulary space and as a safeguard against de-anonymization, phrases that contain vocabulary words with a rarity of $3\sigma$ or above over the whole dataset discarded.

    The original dataset as supplied was provided in HTML text format. However, HTML text annotations (<i>) were stripped easily and formatting tokens (<br/>) were converted to their corresponding string counterparts (\textbackslash n, in this case.)

    As both client questions and psychologist responses often span multiple sentences, each string in the dataset was then chunked into an array of sentences using NLTK \autocite{lopeed2002}. Finally, to match the resulting array of input and output sentences into pairs, sentences in the input/output arrays were grouped in order of their relative distance based on Princeton WordNet \autocite{millgeo1998} with niceties (i.e. ``Hi! Thanks for writing in.'') removed as much as possible based on statics rules on removing stopwords and greetings.

    The resulting dataset contains 2,123 pairs of utterances between the public ``client'' and accredited psychotherapists in text format with a vocabulary of roughly 15,000 tokens.

    For sequence-modeling applications, however, this is a rather small set of data which would not be enough to train a responsive language model. Hence, the Cornell Movie Dataset \autocite{mizicri2011} --- pre-prepared pairwise utterances between film characters standardized for sequence modeling applications --- is used as the base language understanding dataset used to train the network.

    \subsubsection{Data Preparation}

    The resulting utterance pairs from the cleaned dataset were then dependently shuffled within their sub-dataset (that is, movie data is initially not mixed with psychology data.) 

    The dataset is then fully word-tokenized while reserving the tokens $0$, $1$, and $2$ for special use. Each utterance is capped and tailed using the tokens $1$ and $2$, corresponding to <sos> (start of sequence) and <eos> (end of sequence) respectively. Finally, the maximum length of all prepared sequences is taken, and the whole dataset is $0$-padded to match the length of that maximum sequence.

    In addition to padding, however, the output data is also ``right-shifted'' by one space: one extra pad token is prepended to the output sequences, and one is taken away from the end. This step, as documented by \textit{Popel, et al.} \autocite{popemar2018} and \textit{Vaswani, et al.} \autocite{vaswash2017}, will greatly reduce the likelihood of the model learning to simply copy the input sequence directly as the output sequence will need to be, at a minimum, associated with the corresponding right-shifted version via context.

    Finally, the training data is shaped into minibatches of 48 samples each. Special care was taken in mixing the movie dialogue corpus and the counseling corpus. All minibatches were created by randomly selecting samples from either the movie corpus or counseling corpus weighted by on the location of the minibatch in the final dataset. In earlier samples, there is a near-zero chance of the selection of the counseling dataset, but, once 4,000 warm-up batches have passed, both datasets were selected with near equal probability until exhaustion; through training exclusively on the pre-prepared gold-standard dataset (i.e. dialogue corpus) for recently seeded weights, this work aims to build higher quality utterances in the model before training on its specialization. Formally, the probability of the psychotherapy corpus being selected per mini-batch is defined as follows:

   \begin{equ}[H]
        \begin{equation}
            P(b) = \frac{N_0K}{(K-N_0)e^{-rb}+N_0}
        \end{equation}
        \caption{Psychotherapy Data Mixing Probability, where, $N_0=1\times10^{-3}$, $K=5\times10^{-1}$, and $r=2.5\times10^{-3}$}
    \end{equ}

    \medskip

    Notably, this is not strictly transfer training. As this mixing of dataset is done before training takes place, the model is simply learning a larger dataset in conjunction with a smaller one. This helps prevent over-fitting --- whereby the model specializes too much on the smaller target dataset to the point of sheer memorization --- and maintains the state of the optimizer --- which helps accelerate training.

    The resulting dataset is organized into 48-item minibatches of pairs of utterances in the ``prompt-and-answer'' style: the input is a set of utterances by the client/prompter, and the desired output is the target response by the physiotherapist/responder. Approximately $\frac{1}{15}$ of the data consists of psychotherapy replies, and the remaining $\frac{14}{15}$ is the aforementioned gold-standard dialogue corpus. Although the fraction of psychotherapy data is comparatively small due to the lack of high-quality psychotherapy data available for research as aforementioned, the model was able to specialize enough to achieve appreciable results as will be discussed later.

    \subsubsection{Markov Assumption} \label{markov}
    It is important here to recognize that the inherent Markovian nature in the method by which the language model is trained. This mechanism of training by ``prompt-and-answer'' sentence pairs gives little context towards the history or longer-term knowledge that would usually build between a psychotherapist and client. However, as this model is \textit{not} meant to replace psychotherapists and instead is meant to both communize and reduce the load on the point of care, this training target should be adequate for the desired goal. Furthermore, this method has also previously shown promising results in automated psychotherapy applications \autocite{althtim2016}.

    \subsection{Language Model}
    \subsubsection{Model Topology}
    As per aforementioned, the prepared utterances will be modeled using a encoder-decoder transformer discussed in section \ref{transformers}. Such models allow sequence-to-sequence generation of psychotherapist-mimicking text with state-of-the-art results reasonably efficiently.
    
    In the findings of \textit{Popel, et al.} \autocite{popemar2018}, transformers typically perform the best at quite large batch sizes around 1,500 per gradient update and at reasonably long sequence lengths. Furthermore, the work indicates that --- when paired with large batch sizes --- large model sizes typically perform the best in terms of the quality of final convergence.

    \begin{table}[H]
        \centering
        \begin{tabular}{c|c}
            \textbf{Parameter} & \textbf{Value} \\ \hline
            Encoder Blocks & 2 \\
            Decoder Blocks & 2 \\
            Attention Heads & 6 \\
            Minibatch Size & 48 \\
            Effective Batch Size & 1,536 \\
            Gradient Accumulation Size & 32 \\
            In-Attention Feed Forward Size & 512 \\
            In-Network Feed Forward Size & 256 \\
        \end{tabular}
        \caption{Chosen Network Parameters}
        \label{parameters}
    \end{table}

    However, due to the hardware limitations of this experiment, the model size as indicated as optimal by \textit{Popel, et al.} was not able to be matched. But, when possible, the parameter sizes that the aforementioned group found as optimal were selected. The exact parameters chosen are shown in table \ref{parameters}.

    \subsubsection{Model Training} \label{training}
    The base language model is trained over a period of 7 days and 27 minutes over the aforementioned dataset. In total, 2,288 epochs were processed until the model was deemed to converge. 

    For model optimization, the gradient-descent optimizer Adam was used with $\beta_1=0.9$ and $\beta_2=0.98$. Additionally, $\epsilon=1 \times 10^{-9}$. These parameters are consistent with those employed by \textit{Vaswani, et al.} in order to reduce the rate by which Adam adapts the learning rate from the usual defaults.

    The base learning rate of the model was set to $5.7\times 10^{-2}$, with a built-in 4,000 minibatches of warm up and later an inverse-radical decay. The exact function used to determine learning rate is as follows:

    \begin{equ}[H]
        \begin{equation}
            \begin{cases}
                LRF(s) = min(\frac{1}{\sqrt{s+(1\times10^{-8))}}}, s \times (4000^{-1.5})) \\
                LR(s) = LRF(s) \cdot 5.7 \times 10^{-2}
            \end{cases}
        \end{equation}
        \caption{Learning Rate $LR$, where $s$ is the number of training steps elapsed}
    \end{equ}

    The language model was written in PyTorch \autocite{paszadam2019}, using the peer-reviewed implementation of self-attention written by Huggingface, inc. \autocite{wolftho2020} Data preparation and handling (e.g. word tokenization, sentence chunking, etc.) were performed with Python packages Gensim and NLTK using the Princeton WordNet \autocite{millgeo1998} and \textit{U of Penn.} Treebank \autocite{prasras2008} on Python v3.9. 

    Explorative training of the network was accelerated on one NVIDIA Tesla K80 and one NVIDIA RTX3080 GPUs (Graphical Processing Units). Final, reported training trials in this study were trained and benchmarked on one NVIDIA RTX3070 GPU.

    \subsection{Model Evaluation} \label{evaluation}
    Sequence-to-sequence models are traditionally difficult to evaluate. In language modeling implementations (unlike translation, where there may be patterns to be drawn through metrics like BLEU), there is no easily automated benchmark for evaluating a sentence's coherence and effect. This lack of benchmark is partly due to the fact that multiple, drastically different responses could amount to the same semantic effect (e.g ``I am doing well, thank you.'' vs. ``Doing fine! Thanks.'', etc.)

    In order to address this, a two-part evaluation rubric was created corresponding to previously published work \autocite{deriujan2020} \autocite{xiaozia2020if}. The generated prompt/response pairs and their corresponding human-written validation prompt/response pair were both scored upon the devised rubric in a single-blinded fashion. Through this procedure, the model can be benchmarked by comparing the rubric values that the generated responses received against that received by the corresponding human-written validation responses

    As aforementioned, the scoring rubric was created in two parts. 

    First, a single-blinded adaptation of the Turing test named ``Spot The Bot'' as devised by \textit{Deriu, et al.} \autocite{deriujan2020} was administered. In this segment, blinded human evaluators were tasked with scoring each prompt/response pair on a scale between 1 to 3 --- with 1 representing ``Likely Generated'' to 3 representing ``Likely Human-Written.'' The responses were shuffled such that there exists no pattern in the order by which generated vs. human-written responses appear.

    Second, an altered measurement of Response Quality Index (RQI) similar to that proposed by \textit{Xiao, et al.} \autocite{deriujan2020} was collected by scoring each response upon a set of qualitative metrics. The rubric by which human evaluators used to score responses are listed in table \ref{rubric}.

    \begin{table*}[ht]
        \rowcolors{2}{white}{gray!25}
        \centering
        \begin{tabular}{|c|c|c|c|c|}
            \hline
            \textbf{Parameter} & \textbf{1} & \textbf{2} & \textbf{3} & \textbf{4} \\ \hline \hline
            \textit{Clarity} & \specialcell{Unclear,\\Incoherent} & \specialcell{Coherent,\\ Illogical} & Logical & Logical and Clear \\ \hline
            \textit{Specificity} & Irrelevant & \specialcell{Addresses \\ Prompt} & \specialcell{Engages \\ Prompt} & \specialcell{Offers \\ Opinions} \\ \hline
            \textit{Psychotherapeutic Benefit} & \specialcell{Negative\\influence} & \specialcell{Null/Unclear} & \specialcell{Addresses\\need} & \specialcell{Positive\\influence} \\ \hline
        \end{tabular}
        \caption{Standardized Response Quality Index Scoring Rubric, Higher is Better}
        \label{rubric}
    \end{table*}

    The final ``quality factor'' result from this second part of validation is, as is similar to \textit{Xiao, et al.}, the product of the scores from all three evaluated factors; formally:

     \begin{equ}[H]
        \begin{equation}
            RQI = \sum Clarity_i \times Specificity_i \times Benefit_i
        \end{equation}
        \caption{Response Quality Index}
        \label{rqi_equ}
    \end{equ}

    In validation samples from the non psycho-therapeutic dataset, the value of the psychological benefit section is set as 2 for aggregate QI calculations --- that it literally has a null psychotherapeutic benefit --- and is disregarded for individual psychotherapeutic analysis metrics.

    During data analysis, the raw results for the model's $RQI$ and ``Spot the Bot'' performances were reported and benchmarked against the results coded from human responses.

    In total, 134 coded pairs were evaluated in this manner to measure the validation performance of the model.

    \section{Results and Validation}
    As per aforementioned, the proposed language model was trained on the data prepared in the manner as described in section \ref{training} for a total of 2,288 epochs (roughly 4,000,000 minibatches) until visual convergence. The model was evaluated upon 134 hand-coded prompts paired with ``true'' human replies via the procedure highlighted in \ref{evaluation}.

    \begin{figure*}[ht]
        \centering
        \includegraphics[width=0.95\textwidth]{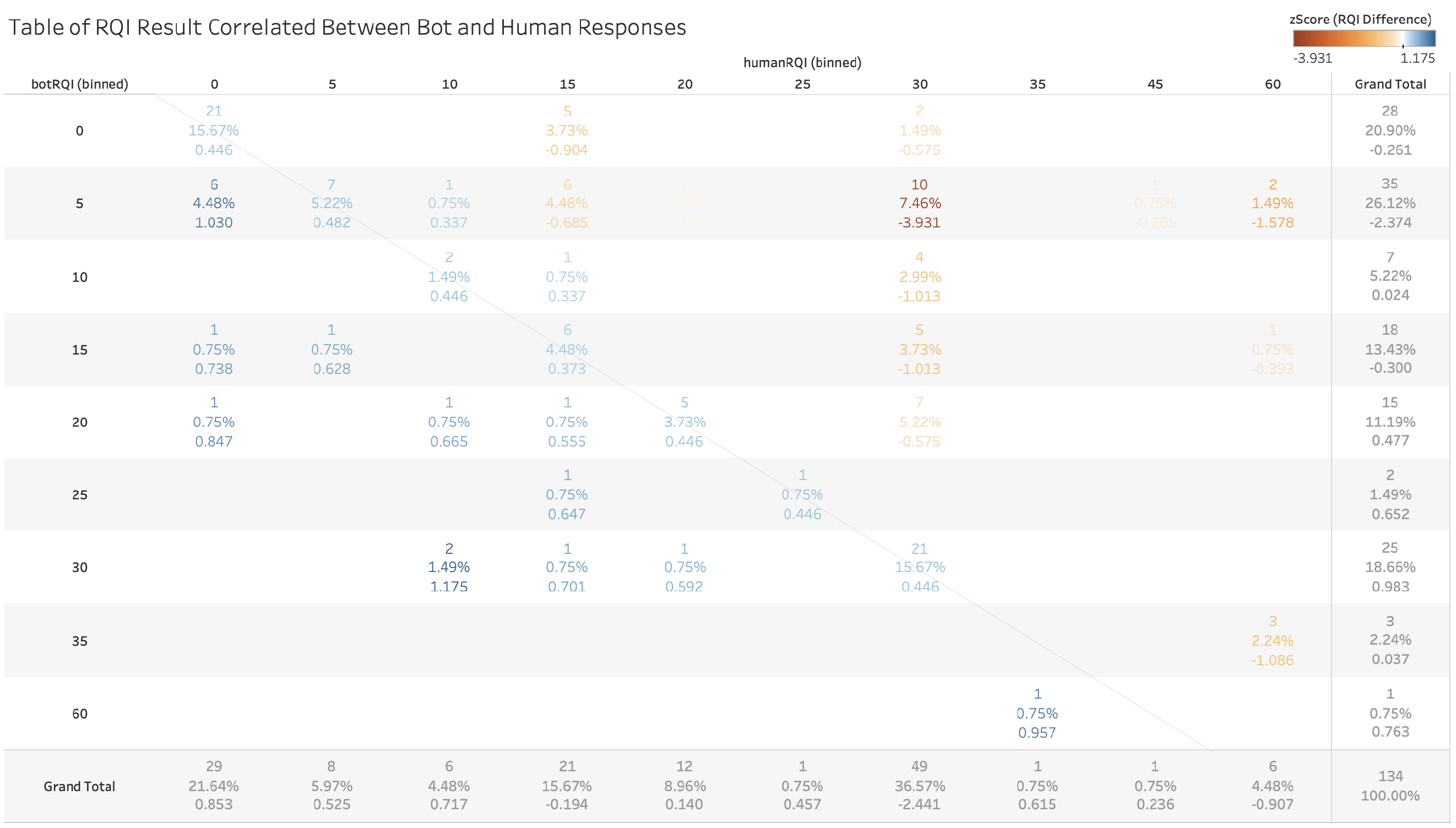}
        \caption{Response Quality Index (RQI) Correlated Between Model and Human Utterances. Each cell reports the count of, percentage of, and z-score of difference between human and machine written utterance that received the pair of RQI scores in question. Center gray line highlights when human and model utterances were scored similarly. Higher values represent more optimal results. \textit{Correlation p<0.001; n=134}}
        \label{rqi_corr}
    \end{figure*}

    \begin{figure*}[ht]
        \centering
        \includegraphics[width=0.95\textwidth]{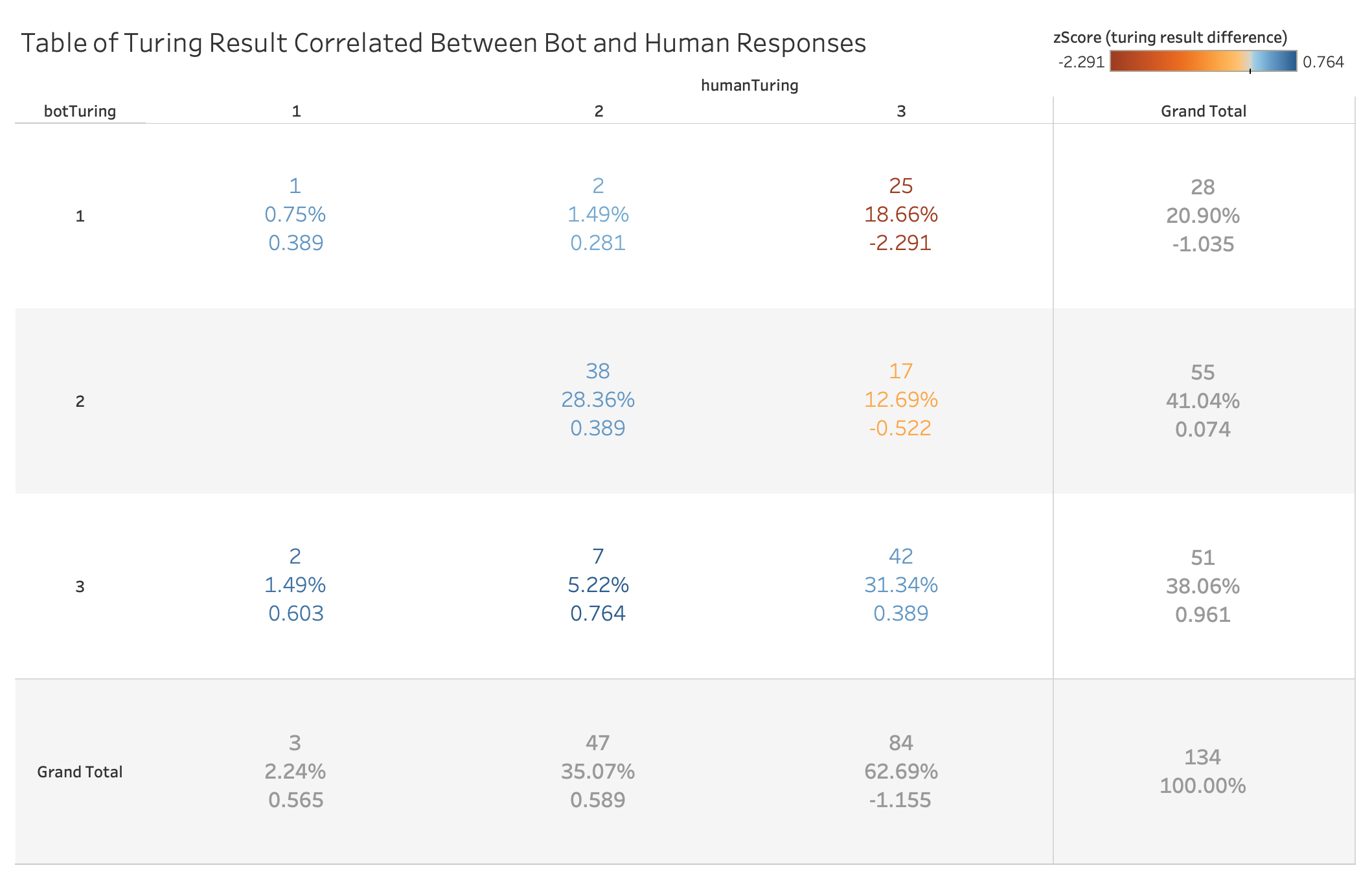}
        \caption{Turing Result Correlated Between Model and Human Utterances. Each cell reports the count of, percentage of, and z-score of difference between human and machine written utterance that received the pair of turing test evaluations in question. Higher values represent a higher semblance of being human-written. \textit{Correlation p<0.001; n=134}}
        \label{turing_corr}
    \end{figure*}

\begin{figure}[H]
        \centering
        \includegraphics[width=0.4\textwidth]{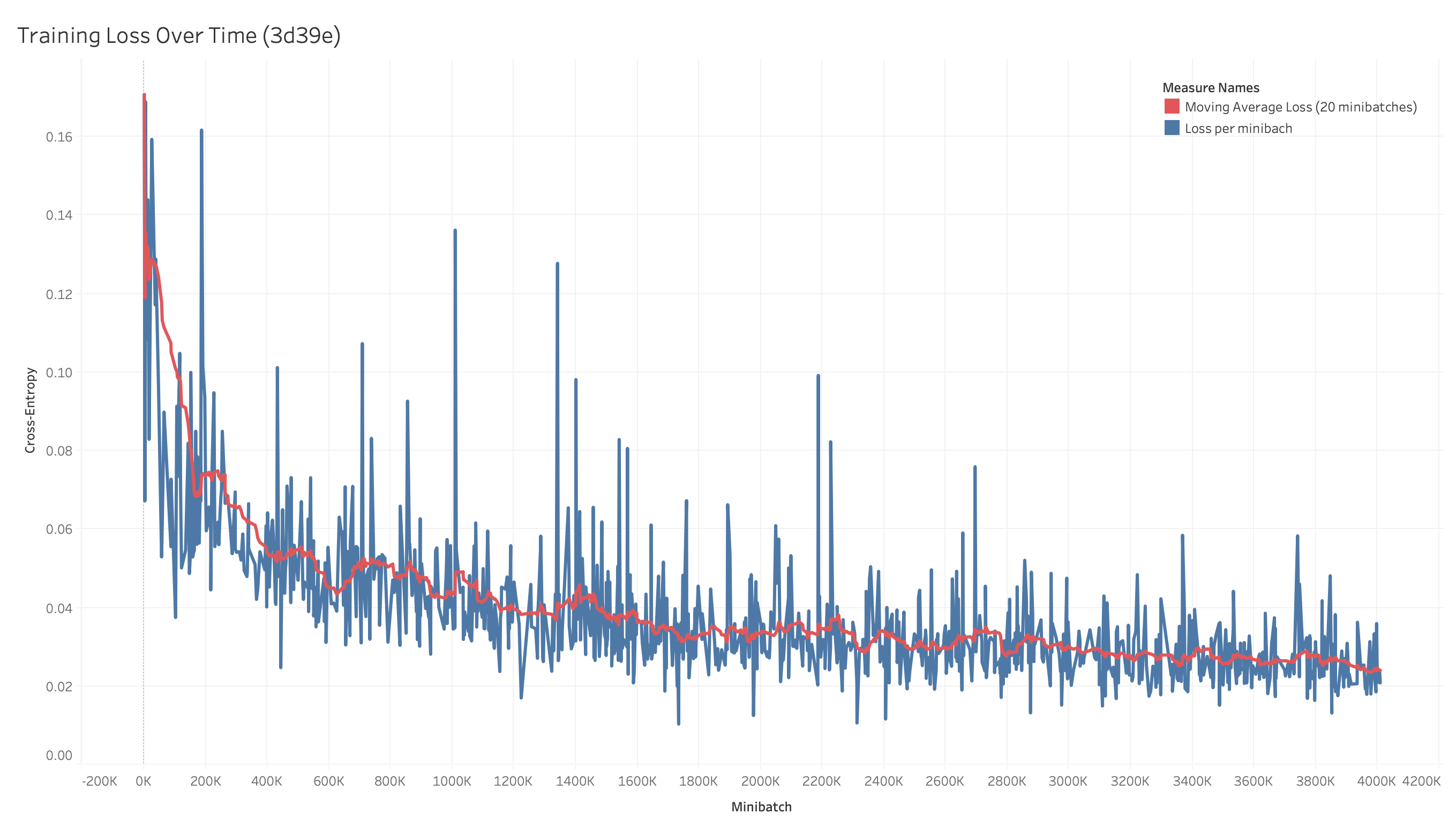}
        \caption{Training Loss per Minibatch}
        \label{loss}
    \end{figure}

    The progression of the training of the final evaluated model is shown in figure \ref{loss}. After roughly 2,800,000 minibatches, the model began to show sign of convergence. When training was stopped, the average rate of change in loss is roughly $-1.42 \times 10^{-8}$ --- a reasonably small value indicating the near-convergence to a local minima.

    After the model's convergence, it was able to achieve reasonably sufficient ability to carry a conversation. An example of such a conversation is shown in dialogue \ref{democonvo}.

    \begin{dia}[H]
        \begin{dialogue}
            \speak{User} I really want to quit smoking, but I am not sure where I should start.
            \speak{Language Model} hopefully you feel you have the time to follow this procedure.
            \speak{U} Yeah, I do! How should I start quitting?
            \speak{L} I have a mental professional near.
            \speak{U} That's a good idea.
            \speak{L} Thank you for your question.
        \end{dialogue}
        \caption{Conversation Demo (Quit Smoking)}
        \label{democonvo}
    \end{dia}

    As highlighted in dialogue \ref{democonvo}, the model was able to synthesize the illusion of continued conversation even with the Markov assumption as discussed in section \ref{markov} in place.

    After training the model to convergence, it subsequently scored against the metrics as highlighted in section \ref{evaluation}. Specifically, this project aims to benchmark the model against previously seen human-written responses of the same prompt --- and analyzing whether the scores given to its utterances meets the scores given to written human responses.

    In total, an evaluation set of 134 human-written sentence pairs were built and benchmarked against the model's responses.

    As shown in figure \ref{rqi_corr}, the model shows promising results when benchmarked for response quality against human responses. $59.7\%$ of the machine-generated responses were coded to be at or above human ``quality'' as per measured by the Response Quality Index shown in equation \ref{rqi_equ}. For the values at which the model did not perform at or better than human-levels, only $17.9\%$ were ``significantly'' deviant ($z>-1\sigma$) in favor of human responses from the mean difference between human and model RQI of 4.1 points.

    Similarly, as shown in figure \ref{turing_corr}, the model was able to meet or exceed human performance for ``Spot the Bot'' \autocite{deriujan2020} Turing evaluations for $67.16\%$ of the validation set. In fact, 3 response data-points even coded for human responses being most likely generated, and in two of which the model achieved a higher score. In this case, the model's utterances was successfully recognized as actually generated for $20.9 \%$ of the sample space. 

    \section{Conclusion and Discussion}
    In this experiment, an automated psychotherapy response mechanism aimed at lessoning the stigma surrounding psychotherapeutic care and easing the volume at human point-of-care was created. Through using state-of-the-art sequence-to-sequence natural language generation (NLG) models --- namely, the encoder/decoder style Transformer --- a language model was made to specialize upon open-source psychotherapist-client text conversations.

    Through training the model, variations upon the original Transformer were made to ease computational overhead. A small, anonymized, declassified psychotherapeutic dataset was acquired through permissive MIT licensing. To mitigate for a lack of high-quality psychotherapeutic data, the Cornell Movie Dialogue dataset was leveraged to create a base language model. A ``trickle-down'' data feeding mechanism was created to slowly intermix the movie dialogue dataset and the much smaller psychotherapeutic dataset.

    After training, the model was evaluated based on variants of two sets of published metrics. Single-blinded, independently coded responses show that the model was able to synthesize utterances at or above human level $59.7\%$ and $67.16\%$ of the time for RQI (response quality index) \autocite{xiaozia2020if} and a comparative Turing test (``Spot the Bot'', \autocite{deriujan2020}) respectively for a test set of 134 validation prompt/response pairs. Of the samples where human responses did outperform synthesized responses, only $17.9\%$ and $20.9\%$ of the subset did so significantly.

    Although the machine generated results did not indicate an outright out-performance of human psychotherapy, the model was able to supply acceptable responses to human psychotherapeutic utterances for most of the dataset. Hence, the results of the model nevertheless represents a major stride towards the original goal of communizing and easing value at psychotherapeutic point-of-care.

    Due to the limitation in dataset, the model was not able to generalize as well as perhaps needed to completely alleviate the work of psychotherapists; however, through a larger dataset, improvements to the language model used, and more training time and resources, it is possible that such services could be fully automated through a mechanism similar to the one proposed. 
\end{multicols}

\pagebreak 
\section{Acknowledgements}
I would like to record my appreciation for the computer science department at The Nueva School. I thank Mr. Albert Huang at The Nueva School for his time and resources during exploratory training, for Mr. Wes Chao, Dr. John Feland, and Dr. Mark Hurwitz at the Nueva School for their support during experimental design, and finally for Mr. Klint Kanopka at Stanford University for his valuable insights and guidance on model design and benchmarking.

\printbibliography
\end{document}